\documentclass{article}
\usepackage{spconf,amsmath,graphicx,cite, subfig, booktabs,url}

\title{BACK-TRANSLATION-STYLE DATA AUGMENTATION FOR END-TO-END ASR}
\name{\scalebox{0.9}[1]{${}^1$Tomoki Hayashi, ${}^2$Shinji Watanabe, ${}^3$Yu Zhang, ${}^1$Tomoki Toda, ${}^4$Takaaki Hori, ${}^5$Ramon Astudillo, ${}^1$Kazuya Takeda}}
\address{${}^1$Nagoya University, Nagoya, JAPAN\\
     ${}^2$Center for Language and Speech Processing, Johns Hopkins University, Baltimore, MD, USA\\
     ${}^3$Google, Inc., USA\\
     ${}^4$Mitsubishi Electric Research Laboratories (MERL), Cambridge MA, USA\\
     ${}^5$Spoken Language Systems Lab, INESC-ID Lisboa, Portugal\\
     }

\begin{document}
%
\maketitle
\begin{abstract}
In this paper we propose a novel data augmentation method for attention-based end-to-end automatic speech recognition (E2E-ASR), utilizing a large amount of text which is not paired with speech signals.
Inspired by the back-translation technique proposed in the field of machine translation, we build a neural text-to-encoder model which predicts a sequence of hidden states extracted by a pre-trained E2E-ASR encoder from a sequence of characters.
By using hidden states as a target instead of acoustic features, it is possible to achieve faster attention learning and reduce computational cost, thanks to sub-sampling in E2E-ASR encoder, also the use of the hidden states can avoid to model speaker dependencies unlike acoustic features.
After training, the text-to-encoder model generates the hidden states from a large amount of unpaired text, then E2E-ASR decoder is retrained using the generated hidden states as additional training data.
Experimental evaluation using LibriSpeech dataset demonstrates that our proposed method achieves improvement of ASR performance and reduces the number of unknown words without the need for paired data.
\end{abstract}
\begin{keywords}
automatic speech recognition, end-to-end, data augmentation, back-translation
\end{keywords}

\section{INTRODUCTION}
\label{sec:intro}
Automatic speech recognition (ASR) is the task of converting a continuous speech signal into a sequence of discrete characters, and is a key technology for the realization of natural interaction between humans and machines.
ASR technology has great potential in various applications such as voice search and voice input, making our lives more convenient.
Typical ASR systems~\cite{jelinek1976continuous} consist of multiple modules such as an acoustic model, a lexicon model, and a language model.
Dividing ASR systems into modules makes it possible to optimize each of them separately, but this also results in more complex systems and imposes performance limitations.
Over the past few decades, this approach has been the basis of ASR systems.

With the improvement of deep learning techniques, end-to-end (E2E) approaches have begun to attract attention~\cite{chorowski2014end}.
While typical ASR systems convert a sequence of acoustic features into text step-by-step using several modules trained separately, E2E-ASR systems directly convert speech using a single neural network.
Therefore, the whole E2E-ASR system can be optimized jointly, making system construction much easier than with typical ASR systems.
Furthermore, it does not require costly lexical information or morphological analysis.

The present E2E-ASR approaches can be divided into two types.
First type is based on connectionist temporal classification (CTC)~\cite{graves2006connectionist,graves2014towards,chorowski2014end,amodei2016deep,soltau2016neural}.
The CTC approach makes it possible to map the input sequences of acoustic features to output sequences of symbols of shorter length without using a hidden Markov model (HMM).
However, it requires assumptions of conditional independence in the output sequence, i.e., each output symbol such as a character or phoneme is independently predicted in each frame.
The second E2E-ASR approach utilizes an attention-based sequence-to-sequence (Seq2Seq) model~\cite{chorowski2015attention}.
In this approach, a sequence of acoustic features is directly mapped into text using an encoder-decoder architecture~\cite{cho2014learning,sutskever2014sequence}.
In contrast to the CTC-based approach, the attention-based Seq2Seq approach is not bound by any assumptions, therefore it can be trained to directly maximize the probability of a word sequence given a sequence of acoustic features.
However, in exchange for its generality, the Seq2Seq approach requires large amounts of data for training.
Furthermore, since the language model is not a separate module, the large amounts of text typically available cannot be used to improve its performance.
This actually yields significant degradation of proper noun recognition, which are not appeared in the paired speech and text data, and affects negatively to production when evaluated on live production data according to \cite{google2017improving}.
\begin{figure*}[t]
\begin{center}
\includegraphics[width=1.55\columnwidth]{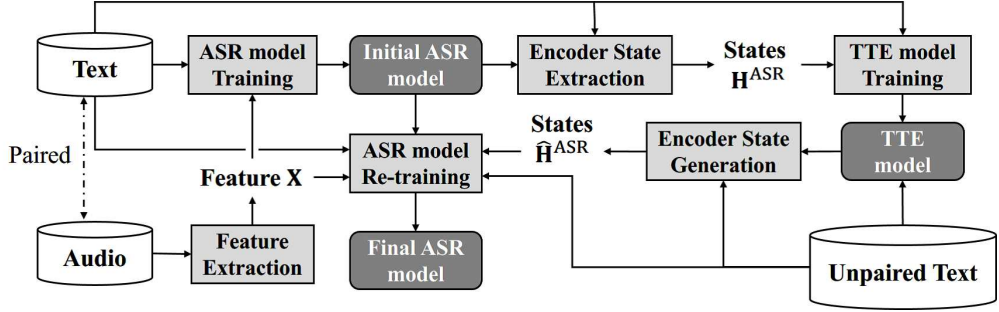}
\end{center}
\vspace{-5mm}
\caption{\it Overview of proposed back-translation-style data augmentation method.}
\vspace{-3mm}
\label{fig:overview}
\end{figure*}

One straightforward approach to address these issues is to integrate a language model with the Seq2Seq model, including shallow fusion, deep fusion, and their variants~\cite{chorowski2016towards,gulcehre2015using,sriram2017cold}.
Shallow fusion~\cite{chorowski2016towards,hori2017advances} is the most simple approach in that we separately train a Seq2Seq model and a language model and then combine the score of two models in the decoding phase.
Deep fusion~\cite{gulcehre2015using} is an approach which has been proposed in the field of neural machine translation.
A seq2seq model and a language model are trained separately, and then the hidden states of the decoder of the Seq2Seq model and those of the language model are concatenated using a gating matrix which controls the importance of each model.
The parameters used to calculate the gating matrix are then trained using a small amount of training data while fixing all of the other parameters.
These fusion approaches enable us to utilize a large amount of unpaired text to improve ASR performance.
The resulting model is not actually end-to-end, however, since it requires additional steps and fine-tuning to integrate the separate modules.

A simpler approach is back-translation~\cite{sennrich2015improving,lample2017unsupervised}, a method which has been proposed in the field of machine translation.
In this approach, a pre-trained target-to-source translation model is used to generate source text from unpaired target text. Augmenting training data with back-translated data led to notable improvements in performance of neural machine translation models\cite{sennrich2015improving}.
Similar techniques have also led to performance improvements in related tasks such as automatic post edition~\cite{junczys2016log}.

Inspired by the back-translation approach, in this paper we propose a novel data augmentation method for attention-based E2E-ASR models allows them to utilize large amounts of text not paired with speech signals.
Instead of using a text-to-speech system on unpaired text to produce synthetic speech~\cite{tjandra2017listening} or using grapheme to phoneme conversion to generate paired text and pseudo speech sequences based on phonemes~\cite{renduchintala2018multi}, we build a text-to-encoder model which learns to predict the hidden states of the E2E-ASR encoder.
Targeting the states of the speech encoder, rather than speech itself makes it possible to achieve faster attention learning and reduce computational cost, thanks to sub-sampling present in E2E-ASR encoder, Furthermore, the use of the hidden states can avoid to model speaker dependencies unlike acoustic features.
After training, the text-to-encoder model generates the hidden states from a large amount of unpaired text, and then the decoder of the E2E-ASR model is retrained using the generated hidden states as additional training data.
To evaluate our proposed method, we conduct experimental evaluation using LibriSpeech dataset~\cite{panayotov2015librispeech}.
The experimental results demonstrate that our proposed method achieves the improvement of ASR performance and makes it possible to improve the recognition results for unknown words.

\section{BACK-TRANSLATION-STYLE \\ DATA AUGMENTATION}
\label{sec:proposed}
\begin{figure*}[t]
\begin{center}
\vspace{0mm}
\subfloat[Attetion-based ASR model]{\includegraphics[width=0.6\columnwidth]{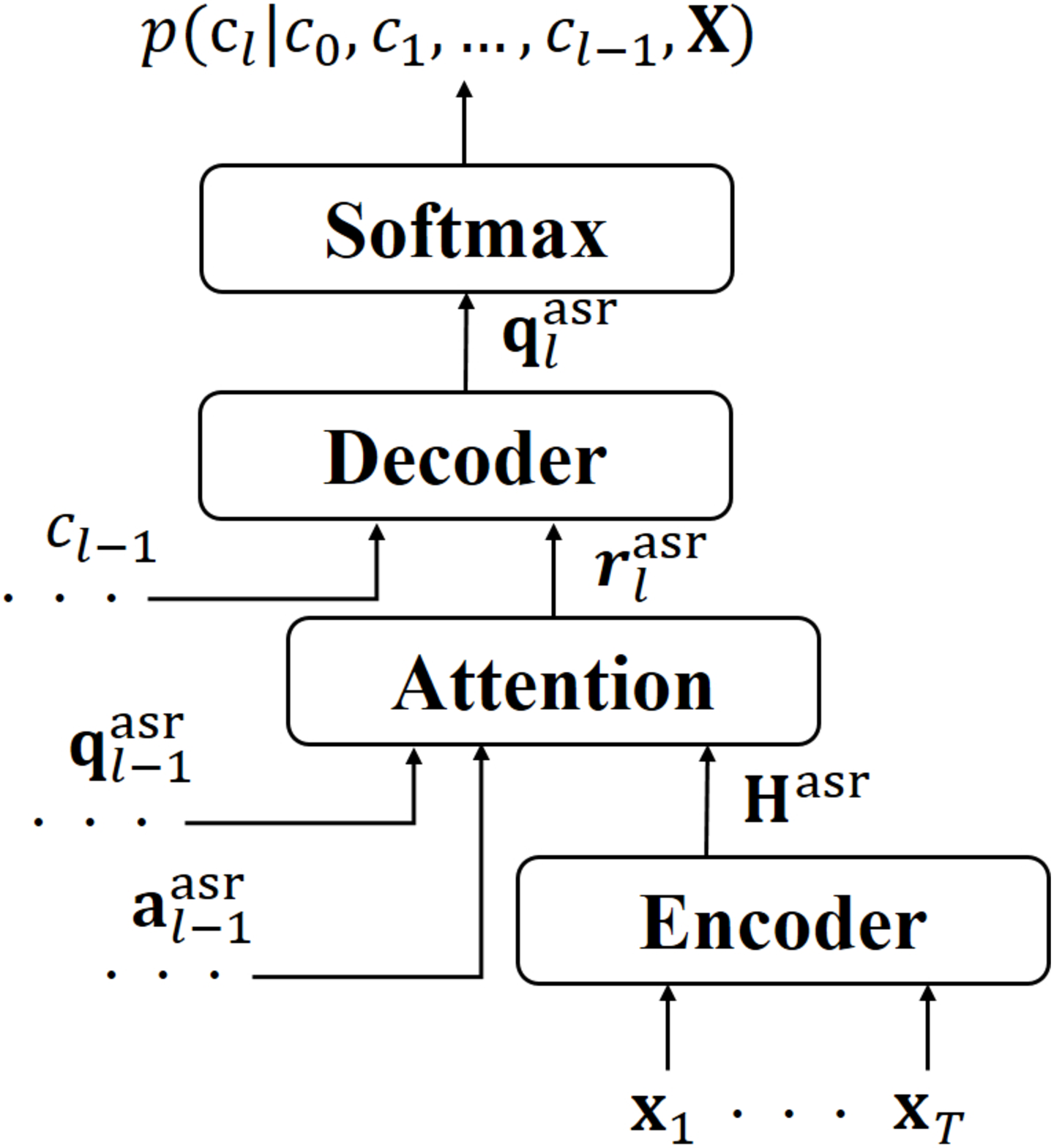}}\quad\quad\quad
\subfloat[Tacotron2-based TTE model]{\includegraphics[width=0.9\columnwidth]{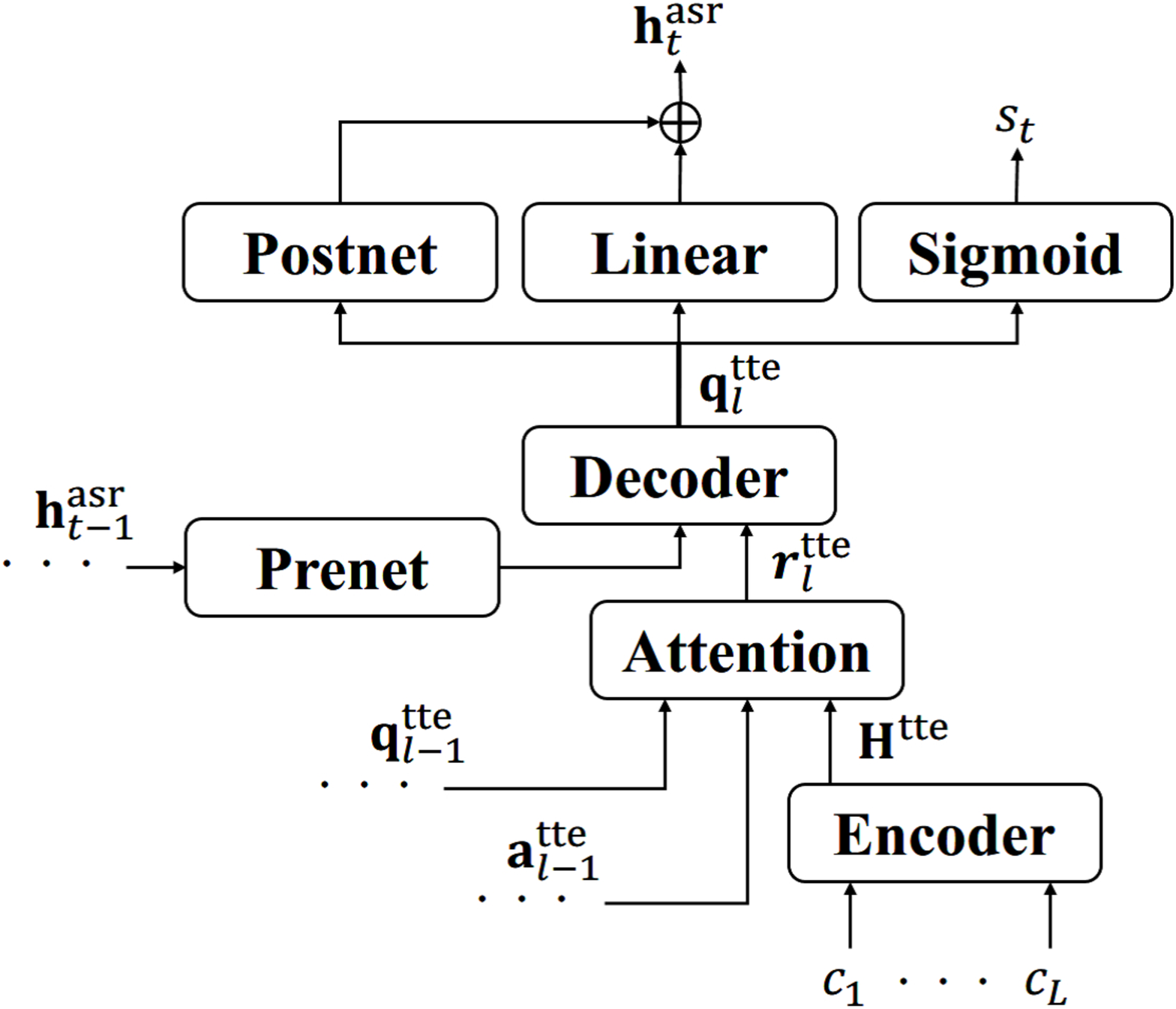}}
\end{center}
\vspace{-4mm}
\caption{\it Overview of attention-based ASR and Tacotron2-based TTE network architectures.}
\label{fig:architecture}
\vspace{-3mm}
\end{figure*}
\subsection{Overview}
\label{ssec:overview}
An overview of our proposed back-translation-style data augmentation method is shown in Fig.~\ref{fig:overview}.
First, the attention-based E2E-ASR model is trained using paired training data which consists of text and speech.
Next, the final layer hidden states of the ASR encoder are extracted, providing paired training data which consists of text and the corresponding hidden states.
Using this paired training data, a neural text-to-encoder (TTE) model is trained to predict the hidden states of the ASR encoder from a sequence of characters.
Finally, the text-to-encoder model generates hidden states from a large amount of unpaired text and the ASR decoder is retrained using the generated states as additional training data.

\subsection{ASR model training}
\label{ssec:att_e2e_asr}
An overview of an attention-based ASR model is shown in Fig.~\ref{fig:architecture}(a).
This model directly estimates posterior $p(\mathbf{C}|\mathbf{X})$, where $\mathbf{X} = \{\mathbf{x}_1,\mathbf{x}_2, \dots, \mathbf{x}_T\}$ represents a sequence of input features, and $\mathbf{C} = \{c_1, c_2, \dots, c_L\}$ represents a sequence of output characters.
Posterior $p(\mathbf{C}|\mathbf{X})$ is factorized with a probabilistic chain rule as follows:
\begin{equation}
    \label{eq:att_posterior}
    p(\mathbf{C}|\mathbf{X}) = \prod_{l=1}^L p(c_l | c_{1:l-1}, \mathbf{X}),
\end{equation}
where $c_{1:l-1}$ represents subsequence $\{c_1, c_2, \dots\, c_{l-1}\}$, and $p(c_l | c_{1:l-1}, \mathbf{X})$ is calculated as follows:
\begin{eqnarray}
    \label{eq:att_encoder}
    \mathbf{h}_t^{\mathrm{asr}} &=& \mathrm{Encoder}^{\mathrm{asr}}(\mathbf{X}), \\
    \label{eq:att_weight1}
    a_{lt}^{\mathrm{asr}} &=& \mathrm{Attention}^{\mathrm{asr}}(\mathbf{q}_{l-1}^{\mathrm{asr}},
                                                  \mathbf{h}_t^{\mathrm{asr}},
                                                  \mathbf{a}_{l-1}^{\mathrm{asr}}),
\end{eqnarray}  
\begin{eqnarray}
    \label{eq:att_weight2}
    \mathbf{r}_l^{\mathrm{asr}} &=& \Sigma_{t=1}^{T} a_{lt}^{\mathrm{asr}}\mathbf{h}_t^{\mathrm{asr}}, \\
    \label{eq:att_decoder}
    \mathbf{q}_l^{\mathrm{asr}} &=& \mathrm{Decoder}^{\mathrm{asr}}(\mathbf{r}_l^{\mathrm{asr}},
                                                                     \mathbf{q}_{l-1}^{\mathrm{asr}},
                                                                     c_{l-1}), \\
    \label{eq:att_posterior}
    p(c_l | c_{1:l-1}, \mathbf{X}) &=& \mathrm{Softmax}(\mathrm{LinB}(\mathbf{q}_l^{\mathrm{asr}})),
\end{eqnarray}
where $a_{lt}^*$ represents an attention weight, $\mathbf{a}_l^*$ represents an attention weight vector (sequence of attention weights $\{a_{l0}^*, a_{l1}^*, \dots, a_{lt}^*\}$), $\mathbf{h}_t^*$ and $\mathbf{q}_l^*$ represent the hidden states of encoder and decoder networks, respectively, $\mathbf{r}_l^*$ represents a letter-wise hidden vector, which is a weighted summarization of the hidden vectors using attention weight vector $\mathbf{a}_l^*$, and $\mathrm{LinB}(\cdot)$ represents a linear layer with a trainable matrix and bias parameters.

All of the above networks are optimized using back-propagation through time (BPTT)~\cite{werbos1990backpropagation} to minimize the following objective function:
\begin{equation}
\begin{split}
    \label{eq:att_obj}
    \mathcal{L}_\mathrm{asr} & = - \log p(\mathbf{C} | \mathbf{X}) \\
                & = - \log\left(\Sigma_{l=1}^L p(c_l | c_{1:l-1}^*, \mathbf{X})\right),
\end{split}
\end{equation}
where $c_{1:l-1}^* = \{c_1^*, c_2^*, \dots, c_{l-1}^*\}$ represents the ground truth of the previous characters.

\begin{figure*}[t]
\begin{center}
\includegraphics[width=1.55\columnwidth]{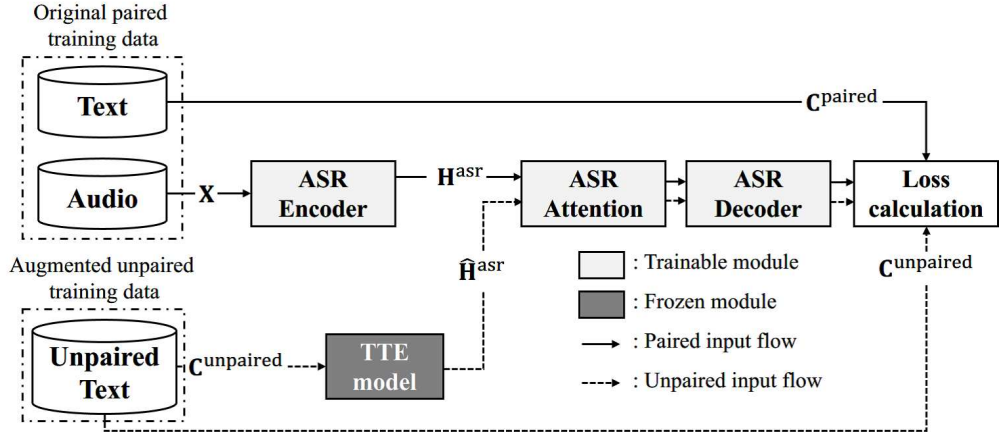}
\end{center}
\vspace{-5mm}
\caption{\it Flowchart of proposed retraining.}
\label{fig:joint_retrain}
\vspace{-3mm}
\end{figure*}

\subsection{TTE model training}
\label{ssec:e2e_tts}
As our neural text-to-encoder (TTE) model we use Tacotron2, which has demonstrated superior performance in the filed of text-to-speech synthesis~\cite{shen2017natural}.
An overview of its network architecture is shown in Fig.~\ref{fig:architecture}(b).
In our framework, the network predicts ASR encoder state $\mathbf{h}_t^{\mathrm{asr}}$ and the probability of the end of sequence $s_t$ at each frame $t$ from a sequence of input characters $\mathbf{C} = \{c_1, c_2, \dots, c_L\}$ as follows:
\begin{eqnarray}
    \label{eq:tts_encoder}
    \mathbf{h}_l^{\mathrm{tte}} &=& \mathrm{Encoder}^{\mathrm{tte}}(\mathbf{C}), \\
    \label{eq:tts_attention}
    a_{tl}^{\mathrm{tte}} &=& \mathrm{Attention}^{\mathrm{tte}}(\mathbf{q}_{t-1}^{\mathrm{tte}},
                                                  \mathbf{h}_l^{\mathrm{tte}},
                                                  \mathbf{a}_{t-1}^{\mathrm{tte}}),
\end{eqnarray}
\begin{eqnarray}
    \label{eq:tts_attention2}
    \mathbf{r}_t^{\mathrm{tte}} &=& \Sigma_{l=1}^{L} a_{tl}^{\mathrm{tte}}\mathbf{h}_l^{\mathrm{tte}}, \\
    \mathbf{v}_{t-1} &=& \mathrm{Prenet}(\mathbf{h}_{t-1}^{\mathrm{asr}}), \\
    \label{eq:tts_decoder}
    \mathbf{q}_t^{\mathrm{tte}} &=& \mathrm{Decoder}^{\mathrm{tte}}(\mathbf{r}_t^{\mathrm{tte}}, \mathbf{q}_{t-1}^{\mathrm{tte}}, \mathbf{v}_{t-1}), \\
    \label{eq:tts_output_before}
    \hat{\mathbf{h}}_t^{b, \mathrm{asr}} &=& \tanh(\mathrm{LinB}(\mathbf{q}_t^{\mathrm{tte}})), \\
    \mathbf{d}_{t} &=& \mathrm{Postnet}(\mathbf{q}_l^{\mathrm{tte}}), \\
	\label{eq:tts_output_after}
    \hat{\mathbf{h}}_t^{a,\mathrm{asr}} &=& \tanh(\mathrm{LinB}(\mathbf{q}_l^{\mathrm{tte}})+ \mathbf{d}_{t}), \\
    \hat{s}_t &=& \mathrm{Sigmoid}(\mathrm{LinB}(\mathbf{q}_t^{\mathrm{tte}})),
\end{eqnarray}
where $\mathrm{Prenet}(\cdot)$ is a shallow feed-forward network to convert the network outputs before feedback to the decoder, $\mathrm{Postnet}(\cdot)$ is a convolutional neural network to refine the network outputs, and $\hat{\mathbf{h}}_t^{b,\mathrm{asr}}$ and $\hat{\mathbf{h}}_t^{a,\mathrm{asr}}$ represent predicted hidden states of the ASR encoder before and after refinement by Postnet.
Note that the indices of the encoder and decoder states are reversed in comparison to the ASR formulation in Eqs.~(\ref{eq:att_encoder})-(\ref{eq:att_posterior}), and that we use an additional activation function $\tanh(\cdot)$ in Eqs.~(\ref{eq:tts_output_before}) and (\ref{eq:tts_output_after}) to avoid mismatching of the ranges of the outputs, in contrast to the  original Tacotron2 architecture~\cite{shen2017natural}.

All of the networks are jointly optimized to minimize the following objective functions:
\begin{equation}
\begin{split}
    \label{eq:tts_obj}
    & \mathcal{L}_\mathrm{tte} = \mathrm{MSE}(\hat{\mathbf{h}}_t^{a,\mathrm{asr}}, \mathbf{h}_t^{\mathrm{asr}}) + \mathrm{MSE}(\hat{\mathbf{h}}_t^{b,\mathrm{asr}}, \mathbf{h}_t^{\mathrm{asr}})\\
    & + \mathrm{L1}(\hat{\mathbf{h}}_t^{a,\mathrm{asr}}, \mathbf{h}_t^{\mathrm{asr}}) + \mathrm{L1}(\hat{\mathbf{h}}_t^{b,\mathrm{asr}}, \mathbf{h}_t^{\mathrm{asr}})\\
    & + \tfrac{1}{T}\Sigma_{t=1}^T s_t \ln \hat{s}_t + (1 - s_t) \ln (1 - \hat{s}_t),
\end{split}
\end{equation}
where $\mathrm{MSE}(\cdot)$ represents mean square error, $\mathrm{L1}(\cdot)$ represent an L1 norm, and the last two terms represent the binary cross entropy for the probability of the end of sequence.
The use of losses for both outputs, before and after Postnet aids fast convergence.
In the original Tacotron2 paper~\cite{shen2017natural}, the L1 norm was not used as the objective function, however, in~\cite{jia2018transfer} it was reported that the use of L1 norm improves performance, especially when using noisy training data.

\subsection{ASR decoder retraining}
\label{ssec:retraining}
After training of the TTE model, we retrain the ASR decoder using both the paired and unpaired training data.
A flowchart of this retraining is shown in Fig.~\ref{fig:joint_retrain}.
We concatenate the paired and unpaired text datasets, and then for each text, if there is paired speech data, the acoustic features of that speech are used as inputs.
If not, the hidden states generated by the TTE model are used as inputs.
Using both the generated hidden states and the original acoustic features produces a regularization effect which prevents overfitting to the generated states.

\section{EXPERIMENTAL EVALUATION}

\label{sec:experiment}
\subsection{Experimental conditions}
We conducted an experimental evaluation using the LibriSpeech dataset~\cite{panayotov2015librispeech}, which consists of two sets of clean speech data (100 hours + 360 hours), and noisy speech data (500 hours) for training.
We used 100 hours of clean speech data to train the initial ASR model and the text-to-encoder (TTE) model, and the text of 360 hours of clean speech data to retrain the ASR decoder.
We used five hours of clean development data as a validation set, and five hours of clean test data as an evaluation set.
To evaluate the effectiveness of our proposed method, we compared the recognition performance of the following seven methods:
\begin{description}
	\setlength{\parskip}{0.5mm}
    \setlength{\itemsep}{0.5mm}
    \item[Baseline]\mbox{}\\
    model trained with 100 hours of acoustic features;
    \item[Retrain-State]\mbox{}\\
    model retrained with 360 hours of generated hidden states and 100 hours of extracted hidden states;
    \item[Retrain-State-Frozen]\mbox{}\\
    model retrained with 360 hours of generated hidden states and 100 hours of extracted hidden states while the attention layers are frozen;
    \item[Retrain-Joint]\mbox{}\\
    model retrained with 360 hours of generated hidden states and 100 hours of acoustic features;
    \item[Oracle-State]\mbox{}\\
    model trained with 460 hours of extracted hidden states;
    \item[Oracle-State-Frozen]\mbox{}\\
    model trained with 460 hours of extracted hidden states while the attention layers are frozen;
    \item[Oracle-Feature]\mbox{}\\
    model trained with 460 hours of acoustic features;
\end{description}
where ``{generated hidden states}'' represent the hidden states generated by the TTE model, and ``{extracted hidden states}'' represent the hidden states extracted from the ASR encoder using raw acoustic features.

We used an acoustic feature vector consisting of an 80-dimensional log Mel-filter bank and three-dimensional pitch features, which were extracted using the open-source speech recognition toolkit Kaldi~\cite{povey2011kaldi}.
The ASR encoder consisted an eight-layered bidirectional long short-term memory with a projection layer~\cite{sak2014long} (BLSTMP), and the ASR decoder consisted a one-layered LSTM\@.
In the second and third layers from the bottom of the ASR encoder, sub-sampling was performed to reduce the length of utterances $T$, yielding the length $T/4$.
The ASR attention network used location-aware attention~\cite{chorowski2015attention}, which is more robust to long sequences than dot-product~\cite{luong2015effective} or additive attention~\cite{bahdanau2014neural}.
For decoding, we used a beam search algorithm~\cite{sutskever2014sequence} with beam size of 20.
We manually set the maximum and minimum lengths of the output sequence to 0.3 and 0.8 times the length of the subsampled input sequence, respectively.
Details of the experimental conditions for the ASR model are shown in Table~\ref{tb:asr_cond}.

\begin{table}[t]
\begin{center}
\caption{\it Experimental conditions for the ASR model.}
\vspace{-2mm}
\label{tb:asr_cond}
\scalebox{0.9}{%
{\renewcommand\arraystretch{0.8}
\begin{tabular}{l l} \toprule
\ Encoder type                   & \ BLSTMP \\
\ \# encoder layers              & \ 8 \\
\ \# encoder units               & \ 320 \\
\ \# projection units            & \ 320 \\
\ Decoder type                   & \ LSTM \\
\ \# decoder layers              & \ 1 \\
\ \# decoder units               & \ 320 \\
\ \# dimension in attention      & \ 300 \\
\ \# filter in attention         & \ 10 \\
\ Filter size in attention       & \ 100 \\
\ Learning rate                  & \ 1.0 \\
\ Gradient clipping norm         & \ 5  \\
\ Batch size                     & \ 50 \\
\ Maximum epoch                  & \ 30 (for initial training)  \\
                                 & \ 15 (for retraining)  \\
\ Optimization method            & \ AdaDelta~\cite{zeiler2012adadelta} \\
\ AdaDelta $\rho$                & \ 0.95 \\
\ AdaDelta $\epsilon$            & \ $10^{-8}$ \\
\ AdaDelta $\epsilon$ decay rate & \ $10^{-2}$ \\
\ Beam size                      & \ 20 \\
\ Maximum length                 & \ 0.8 \\
\ Minimum length                 & \ 0.3 \\ \bottomrule
\end{tabular}
}}
\vspace{-5mm}
\end{center}
\end{table}

\begin{table}[t]
\begin{center}
\caption{\it Experimental conditions for the TTE model.}
\vspace{-2mm}
\label{tb:tts_cond}
\scalebox{0.9}{%
{\renewcommand\arraystretch{0.8}
\begin{tabular}{l l} \toprule
\ Encoder type                   & \ CNN + BLSTM \\
\ \# embedding dimension         & \ 512 \\
\ \# encoder layers              & \ 3 (CNN) \\
                                 & \ 1 (BLSTM) \\
\ \# encoder BLSTM units         & \ 512 \\
\ \# encoder CNN filters         & \ 512 \\
\ Encoder CNN filter size        & \ 5 \\
\ Decoder type                   & \ LSTM \\
\ \# decoder layers              & \ 2 \\
\ \# decoder units               & \ 1024 \\
\ \# dimension in attention      & \ 128 \\
\ \# filters in attention        & \ 32 \\
\ Filter size in attention       & \ 31 \\
\ \# Prenet layers               & \ 2 \\
\ \# Prenet units                & \ 256 \\
\ \# Postnet layers              & \ 5 \\
\ \# Postnet filters             & \ 512 \\
\ Postnet filter size            & \ 5 \\
\ Dropout rate                   & \ 0.5 \\
\ Zoneout rate                   & \ 0.1 \\
\ Learning rate                  & \ $10^{-3}$ \\
\ Gradient clipping norm         & \ 1  \\
\ Batch size                     & \ 50 \\
\ Maximum epoch                  & \ 100 \\
\ Optimization method            & \ Adam~\cite{kingma2014adam} \\
\ Adam $\epsilon$                & \ $10^{-6}$ \\
\ Threshold to stop generation   & \ 0.75 \\ \bottomrule
\end{tabular}
}}
\vspace{-4mm}
\end{center}
\end{table}

\begin{table}[t]
\begin{center}
\vspace{-1mm}
\caption{\it MSE of the TTE model for validation set.}
\label{tb:tts_results}
\vspace{-1mm}
\scalebox{0.9}{%
{\renewcommand\arraystretch{0.8}
\begin{tabular}{l c}
\toprule
\ With L1 norm         & \underline{{\bf 0.0298}} \\
\ Without L1 norm \hspace{8mm}  & 0.0335 \\
\bottomrule
\end{tabular}
}}
\vspace{-5mm}
\end{center}
\end{table}

\begin{table}[b]
\begin{center}
\vspace{-4mm}
\caption{\it ASR performance using various retraining methods.}
\label{tb:asr_results}
\vspace{-2mm}
\scalebox{0.9}{%
{\renewcommand\arraystretch{0.8}
\begin{tabular}{l c c c}
                                & \multicolumn{2}{c}{CER / WER [\%]} \\ \cmidrule{2-3}
                                & Validation   & Evaluation  \\ \toprule
Baseline                        & 11.2 / 24.9  & 11.1 / 25.2 \\
Retrain-State                   & 12.0 / 27.6  & 12.4 / 28.3 \\
Retrain-State-Frozen            & 11.9 / 27.1  & 11.9 / 27.6 \\
Retrain-Joint                   & \underline{\bf 10.3} / \underline{\bf 23.5}  & \underline{\bf 10.3} / \underline{\bf 23.6} \\ \midrule
Oracle-State                    & 7.6 / 16.8   & 8.7 / 18.4  \\
Oracle-State-Frozen             & 8.1 / 18.9   & 8.5 / 19.6  \\
Oracle-Feature                  & 4.7 / 11.4   & 4.6 / 11.8  \\ \bottomrule
\end{tabular}
}}
\vspace{-4mm}
\end{center}
\end{table}

The architecture of the TTE model followed the original Tacotron2 settings~\cite{shen2017natural}.
The input characters were converted into 512-dimensional character embeddings.
The TTE encoder consisted of a three-layered 1D convolutional neural network (CNN) containing 512 filters with the shape 5, a batch normalization and rectified linear unit (ReLU) activation function, and an one-layered BLSTM with 512 units (256 units for forward processing, the rest for backward processing).
Although the attention mechanism of the TTE model was based on location-aware attention~\cite{chorowski2015attention}, we additionally cumulated the attention weight feedback to next step to accelerate attention learning.
The TTE decoder consisted of a two-layered LSTM with 1024 units.
Prenet was a two-layered feed forward network with 256 units and ReLU\@.
Postnet was a five-layered CNN containing 512 filters with the shape 5, a batch normalization, and tanh activation function except in the final layer.
Dropout~\cite{srivastava2014dropout} with a probability of 0.5 was applied to all of the  convolution and Prenet layers.
Zoneout~\cite{krueger2016zoneout} with a probability of 0.1 was applied to the decoder LSTM\@.
During generation, we applied dropout to Prenet in the same manner as in~\cite{shen2017natural}, and set the threshold value of the probability of the end of sequence at 0.75 to prevent from cutting off the end of the input sequence.
Details of the experimental conditions for the TTE model are shown in Table.~\ref{tb:tts_cond}.

All of the networks were trained using the end-to-end speech processing toolkit ESPnet~\cite{watanabe2018espnet} with a single GPU (Titan X pascal).
Character error rate (CER) and word error rate (WER) were used as metrics.

\begin{figure*}[t]
\begin{center}
\vspace{-1mm}
\includegraphics[width=1.5\columnwidth]{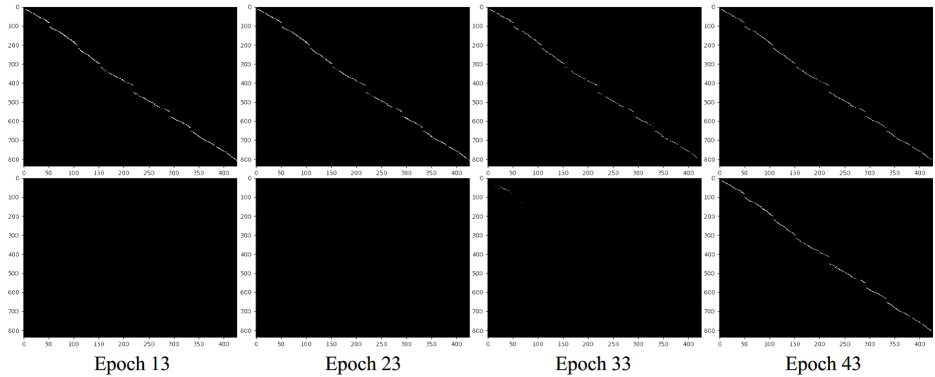}
\end{center}
\vspace{-6mm}
\caption{\it Visualization of attention weight for validation data. Upper figures are w/ L1 norm, bottom ones are w/o L1 norm.}
\label{fig:att_vis}
\vspace{-1mm}
\end{figure*}

\begin{table*}[t]
\begin{center}
\vspace{0mm}
\caption{\it Some recognition examples including unknown words before retraining.}
\label{tb:example}
\vspace{-2mm}
\scalebox{0.9}{%
{\renewcommand\arraystretch{0.8}
\begin{tabular}{l l}
\toprule
GT  & \dots in a reclining attitude being RIGIDLY bound both hands and feet by strong and painful withes \\
ORG & \dots in a reclining attitude being RIGILLY bound both hands and feet by strong and painful with \\
RET & \dots in a reclining attitude being RIGIDLY bound both hands and feet by strong and painful withs \\
\midrule
GT  & the first of our vague but INDUBITABLE data is that there is knowledge of the past \\
ORG & the first of our vague but INDUPINABLE data as that there is knowledge of the past \\
RET & the first of our vague but INDUBITABLE data it set  there is knowledge of the past \\
\midrule
GT  & \dots children that ran about and prattled  when they were in the woods looking  for wild STRAWBERRIES \\
ORG & \dots children that ran about and pratelled when they were in the wood  slooking for wild STRAWBERRES \\
RET & \dots children that ran about and prattled  when they were in the woods looking  for wild STRAWBERRIES \\
\bottomrule
\end{tabular}
}}
\vspace{-4mm}
\end{center}
\end{table*}

\begin{table}[t]
\begin{center}
\vspace{-2mm}
\caption{\it ASR performance of shallow fusion with LM.}
\label{tb:asr_results_lm}
\vspace{-2mm}
\scalebox{0.9}{%
{\renewcommand\arraystretch{0.8}
\begin{tabular}{l c c c}
                      & \multicolumn{2}{c}{CER / WER [\%]} \\ \cmidrule{2-3}
                      & Validation   & Evaluation  \\ \toprule
Baseline + LM         & 10.6 / 23.0  & 10.4 / 22.9 \\
Retrain-Joint + LM    & \underline{\bf 9.8} / \underline{\bf 21.6}  & \underline{\bf 10.0} / \underline{\bf 22.0} \\ \bottomrule
\end{tabular}
}}
\vspace{-4mm}
\end{center}
\end{table}

\subsection{Experimental results}
First, we focus on the effectiveness of adding the L1 norm to the objective function of the TTE model.
Mean square error loss for validation data with teacher forcing is shown in Table~\ref{tb:tts_results}.
We can confirm that the use of the L1 norm results in improved performance.
Furthermore, we found that use of the L1 norm also leads to much faster attention learning.
The attention weights for the validation data are shown in Fig.~\ref{fig:att_vis}.
While the TTE model without the L1 norm is unable to learn the attention until after epoch 40, use of the L1 norm make the model to learn the attention in less than 1/3 the number of epochs.
This is because use of the L1 norm makes the model focus on reducing smaller error, which prevents the decoder of the model from becoming something like an auto-encoder.

Next, we focus on the effectiveness of our proposed data augmentation method.
Our experimental results are shown in Table~\ref{tb:asr_results}.
Compared with the baseline, we can confirm that our proposed ``Retrain-Joint'' approach improved the recognition performance.
However, when only hidden states were used during retraining, no improvement was observed.
This is because using only the hidden states resulted in overfitting.
However, in comparison to the oracle results, we can see that there is a still room for improvement.
These results imply that the use of data of various speakers is more important than the use of various text.
Since hidden states contains less information about speaker characteristics than acoustic features, using hidden states at the targets of the TTE model likely results in the generated hidden states representing the characteristics of an intermediate speaker.
As a result, there is not enough speaker variation among the generated hidden states, degrading the effectiveness of data augmentation.
To address this issue, we will extend our scheme using multi-speaker Tacotron2~\cite{jia2018transfer} in future work.

Some recognition examples including unknown words before retraining are shown in Table~\ref{tb:example}.
We can see that our proposed data augmentation method can improve the performance of the ASR decoder as language model, making it possible to extend the vocabulary. 
A similar effect was observed in the original back-translation work~\cite{sennrich2015improving}.

Finally, the results of shallow fusion with a character-based language model (LM)~\cite{hori2017advances} are shown in Table~\ref{tb:asr_results_lm}, where the LM was trained using text of 360 hours of clean speech, and the balancing weight parameter between two models was decided to achieve the best recognition performance on CER\@.
We can see that the use of LM improved the recognition performance in both cases, indicating that our proposed method can still be combined with LM integration methods.


\section{CONCLUSION}
\label{sec:conclusion}
In this paper we proposed a novel data augmentation method for attention-based E2E-ASR, utilizing a large amount of text which was not paired with speech signals, an approach inspired by the back-translation technique has been proposed in the field of machine translation.
We built a neural text-to-encoder (TTE) model which predicted a sequence of hidden states extracted by a pre-trained E2E-ASR decoder from a sequence of characters.
Using the hidden states as targets makes it possible to achieve faster attention learning and reduces computational cost thanks to sub-sampling in the ASR encoder.
After training, the TTE model generated the hidden states from a large amount of unpaired text, and then the decoder of the E2E-ASR model was retrained using the generated hidden states as additional training data.
An experimental evaluation using LibriSpeech dataset demonstrated that our proposed method achieved the improvement of ASR performance and made it possible to improve the recognition results due to the smaller number of unknown words.
Furthermore, we could confirm that our proposed method can be combined with language model integration methods.

In future works, we will extend the text-to-encoder model to multi-speaker model using speaker embedding vector~\cite{jia2018transfer} to generate more variable hidden states, and apply our proposed method using much larger amount of unpaired text.

\bibliographystyle{IEEEbib}
\bibliography{refs}

\end{document}